\documentclass{article} 
\usepackage{iclr2019_conference,times}

\usepackage{amsmath,amsfonts,bm}

\def\eqref#1{equation~\ref{#1}}

\def\1{\bm{1}}

\DeclareMathAlphabet{\mathsfit}{\encodingdefault}{\sfdefault}{m}{sl}
\SetMathAlphabet{\mathsfit}{bold}{\encodingdefault}{\sfdefault}{bx}{n}

\usepackage{hyperref}
\usepackage{url}

\usepackage{tikz}
\usepackage{bbm} 
\usepackage{scalerel} 
\usepackage{framed} 
\usepackage{amsmath}
\usepackage{subcaption}
\usepackage{forloop}

\usetikzlibrary{shapes,positioning,calc,arrows.meta}
\tikzstyle{lit}=[]
\tikzstyle{clause}=[]
\tikzstyle{lc}=[]
\tikzstyle{ll}=[dashed]
\tikzstyle{send}=[-{Latex[length=1mm, width=2.2mm]}]
\tikzstyle{sendll}=[densely dashed, -{Latex[length=0.72mm, width=1.6mm]}, bend right=10]
\tikzstyle{pcloud}=[cloud, draw=black, aspect=1]
\tikzstyle{mcloud}=[cloud, draw=black, aspect=2.5]
\tikzstyle{boxed}=[ellipse, draw=black]
\tikzstyle{bhead}=[-{Latex[length=1mm, width=2.2mm]}]

\newcommand\plq[1]{} 
\newcommand\dd[1]{}

\newcommand\dhsq[1]{} 

\newcommand\dskip[1]{}

\newcommand\sat[1]{\{#1\}}
\newcommand\mytop{1}
\newcommand\mybot{0}
\newcommand\mean{\mathrm{mean}}

\newcommand\Flip{\mathrm{Flip}}
\newcommand\U{\mathbf{U}}

\newcounter{inum}

\newcommand\SRC{\mathbf{SRC}}
\newcommand\SR{\mathbf{SR}}
\newcommand\PP{\mathbf{PP}}

\newcommand\myNP{\mathbf{NP}}

\newcommand\flip[1]{\overline{#1}}
\newcommand\ie{\emph{i.e.}\ }
\newcommand\eg{\emph{e.g.}\ }
\newcommand\ind[1]{\mathbbm{1}\left\{ {#1} \right\}}
\newcommand\uscore[2]{\mathbf{#1}_{\text{#2}}}

\title{Learning a SAT Solver from Single-Bit Supervision}

\author{Daniel Selsam, Matthew Lamm, Benedikt B\"{u}nz, Percy Liang, David L. Dill \\
  Department of Computer Science \\
  Stanford University \\
  Stanford, CA 94305 \\
  \texttt{\{dselsam,mlamm,buenz,pliang,dill\}@cs.stanford.edu} \\
  \And
  Leonardo de Moura \\
  Microsoft Research \\
  Redmond, WA 98052 \\
  \texttt{leonardo@microsoft.com}
}

\iclrfinalcopy 
\begin{document}

\maketitle

\begin{abstract}
We present NeuroSAT, a message passing neural network that learns to
solve SAT problems after only being trained as a classifier to predict
satisfiability.  Although it is not competitive with state-of-the-art
SAT solvers, NeuroSAT can solve problems that are substantially larger
and more difficult than it ever saw during training by simply running
for more iterations. Moreover, NeuroSAT generalizes to novel
distributions; after training only on random SAT problems, at test
time it can solve SAT problems encoding graph coloring, clique
detection, dominating set, and vertex cover problems, all on a range
of distributions over small random graphs.
\end{abstract}

\section{Introduction}
\label{sec:intro}

The propositional satisfiability problem (SAT) is one of the most
fundamental problems of computer science.
\citet{cook1971complexity} showed that the problem is
$\myNP$-complete, which means that searching for any kind of
efficiently-checkable certificate in any context can be reduced to
finding a satisfying assignment of a propositional formula.
In practice, search problems arising from
a wide range of domains such as hardware and software verification,
test pattern generation, planning, scheduling, and combinatorics are
all routinely solved by constructing an appropriate SAT problem and
then calling a SAT solver~\citep{gomes2008satisfiability}.  Modern SAT
solvers based on backtracking search are extremely well-engineered and
have been able to solve problems of practical interest with millions
of variables~\citep{biere2009conflict}.

\begin{figure}[ht]
  \begin{center}
    \begin{subfigure}{0.35\textwidth}
      \begin{center}
        \[ \left\{ \begin{array}{ll}
          \text{Input:} &  \text{SAT problem } $P$ \\
          \text{Output:} & \ind{P \text{ is satisfiable}}
        \end{array} \right\} \]
      \end{center}
      \caption{Train}
      \hspace{40pt}
    \end{subfigure}
    \begin{subfigure}{0.5\textwidth}
        \begin{center}
          \begin{tikzpicture}[node distance=1cm, auto, draw=black]
            \node[pcloud] (g3) {$P$};
            \node[right=1.0cm of g3, draw=black] (votes3) {NeuroSAT};
            \node[right=1.0cm of votes3] (g3o) {Unsatisfiable};

            \draw[bhead] (g3) to (votes3) ;
            \draw[bhead] (votes3) to (g3o) ;

            \node[pcloud, below=0.05cm of g3] (g4) {$P$};
            \node[right=1.0cm of g4, draw=black] (votes4) {NeuroSAT};
            \node[mcloud, right=1.0cm of votes4, draw=red, color=red] (g4o) {Solution};

            \draw[bhead] (g4) to (votes4) ;
            \draw[bhead, dashed, color=red] (votes4) to (g4o) ;
          \end{tikzpicture}
        \end{center}
        \caption{Test}
      \end{subfigure}
    \end{center}
\caption{We train NeuroSAT to predict whether SAT problems are
  satisfiable, providing only a single bit of supervision for each
  problem. At test time, when NeuroSAT predicts \emph{satisfiable}, we
  can almost always extract a satisfying assignment from the network's
  activations.
  The problems at test time can also be substantially larger, more difficult, and even from
  entirely different domains than the problems seen during training.}
\label{fig:intro1}
\end{figure}
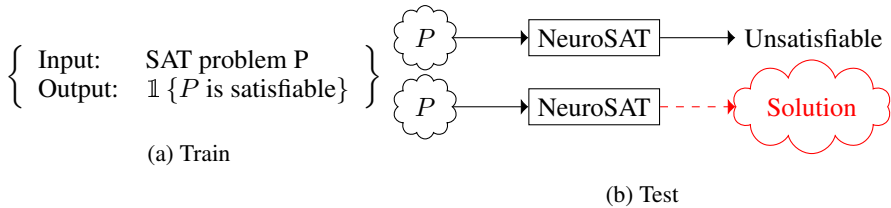

We consider the question: \emph{can a neural network learn to solve
  SAT problems?}  To answer, we develop a novel message passing neural
network (MPNN)~\citep{scarselli2009graph, li2015gated, gilmer2017neural}, \emph{NeuroSAT}, and train it as a classifier to predict
satisfiability on a dataset of random SAT problems. We provide
NeuroSAT with only a single bit of supervision for each SAT problem
that indicates whether or not the problem is satisfiable.  When making
a prediction about a new SAT problem, we find that NeuroSAT guesses
\emph{unsatisfiable} with low confidence until it finds a solution, at
which point it converges and guesses \emph{satisfiable} with very high
confidence. The solution itself can almost always be automatically
decoded from the network's activations, making NeuroSAT an
end-to-end SAT solver.  See Figure~\ref{fig:intro1} for an
illustration of the train and test regimes.

Although it is not competitive with state-of-the-art SAT solvers,
NeuroSAT can solve SAT problems that are substantially larger and more
difficult than it ever saw during training by simply performing more
iterations of message passing.  Despite only running for a few dozen
iterations during training, at test time NeuroSAT continues to find
solutions to harder problems after hundreds and even thousands of
iterations.  The learning process has yielded not a traditional
classifier but rather a procedure that can be run indefinitely to
search for solutions to problems of varying difficulty.

Moreover, NeuroSAT generalizes to entirely new domains.  Since
NeuroSAT operates on SAT problems and since SAT is $\myNP$-complete,
NeuroSAT can be queried on SAT problems encoding any kind of search
problem for which certificates can be checked in polynomial time.
Although we train it using only problems from a single random problem
generator, at test time it can solve SAT problems encoding graph
coloring, clique detection, dominating set, and vertex cover problems,
all on a range of distributions over small random graphs.

The same neural network architecture can also be used to help
construct proofs for unsatisfiable problems. When we train it on a
different dataset in which every unsatisfiable problem contains a
small contradiction (call this trained model \emph{NeuroUNSAT}), it
learns to detect these contradictions instead of searching for
satisfying assignments. Just as we can extract solutions from
NeuroSAT's activations, we can extract the variables involved in the
contradiction from NeuroUNSAT's activations. When the number of
variables involved in the contradiction is small relative to the total
number of variables, knowing which variables are involved in the
contradiction can enable constructing a resolution proof more
efficiently.

\section{Problem Setup}

\emph{Background.} A formula of propositional logic is a boolean
expression built using the constants true ($\mytop$) and false
($\mybot$), variables, negations, conjunctions, and disjunctions.  A
formula is \emph{satisfiable} provided there exists an assignment of
boolean values to its variables such that the formula evaluates to
$\mytop$.  For example, the formula $(x_1 \vee x_2 \vee x_3) \wedge
\neg (x_1 \wedge x_2 \wedge x_3)$ is satisfiable because it will
evaluate to $\mytop$ under every assignment that does not map $x_1$,
$x_2$ and $x_3$ to the same value.  For every formula, there exists an
equisatisfiable formula in \emph{conjunctive normal form} (CNF),
expressed as a conjunction of disjunctions of (possibly negated)
variables.\footnote{This transformation can be done in linear time
  such that the size of the resulting formula has only grown linearly
  with respect to the original formula~\citep{tseitin1968complexity}.}
Each conjunct of a formula in CNF is called a \emph{clause}, and each
(possibly negated) variable within a clause is called a
\emph{literal}. The formula above is equivalent to the CNF formula
$(x_1 \vee x_2 \vee x_3) \wedge (\neg x_1 \vee \neg x_2 \vee \neg
x_3)$, which we can represent more concisely as $\sat{1|2|3,
  \flip{1}|\flip{2}|\flip{3}}$.  A formula in CNF has a satisfying
assignment if and only if it has an assignment such that every clause
has at least one literal mapped to $\mytop$. A \emph{SAT problem} is a
formula in CNF, where the goal is to determine if the formula is
satisfiable, and if so, to produce a satisfying assignment of truth
values to variables. We use $n$ to denote the number of of variables
in a SAT problem, and $m$ to denote the number of clauses.

\emph{Classification task.}  For a SAT problem $P$, we define
$\phi(P)$ to be true if and only if $P$ is satisfiable. Our first goal
is to learn a classifier that approximates $\phi$.  Given a
distribution $\Psi$ over SAT problems, we can construct datasets
$\uscore{\mathcal{D}}{train}$ and $\uscore{\mathcal{D}}{test}$ with
examples of the form $(P, \phi(P))$ by sampling problems $P \sim \Psi$
and computing $\phi(P)$ using an existing SAT solver. At test time, we
get only the problem $P$ and the goal is to predict $\phi(P)$, \ie to
determine if $P$ is satisfiable. Ultimately we care about the
\emph{solving task}, which also includes finding solutions to
satisfiable problems.

\section{Model}
\label{sec:model}
A SAT problem has a simple syntactic structure and therefore could be encoded
into a vector space using standard methods such as an RNN. However,
the semantics of propositional logic induce rich invariances that such
a syntactic method would ignore, such as permutation invariance and
negation invariance. Specifically, the satisfiability of a formula is
not affected by permuting the variables (\eg swapping $x_1$ and $x_2$
throughout the formula), by permuting the clauses (\eg swapping the
first clause with the second clause), or by permuting the literals
within a clause (\eg replacing the clause $1|\flip{2}$ with
$\flip{2}|1$. The satisfiability of a formula is also not affected by
negating every literal corresponding to a given variable (\eg negating
all occurrences of $x_1$ in $\sat{1|\flip{2},\flip{1}|\flip{3}}$ to yield
$\sat{\flip{1}|\flip{2},1|\flip{3}}$).

\begin{figure}[ht]
  \begin{center}
    \begin{subfigure}[t]{0.45\textwidth}
      \begin{center}
\begin{tikzpicture}[node distance=1cm, auto,]
  \node[lit] (x1) {$x_1$};
  \node[lit, right=1.0cm of x1] (nx1) {$\flip{x_1}$};
  \node[lit, right=1.5cm of nx1] (x2) {$x_2$};
  \node[lit, right=1.0cm of x2] (nx2) {$\flip{x_2}$};
  \node[clause, below=0.3cm of nx1] (c1) {$c_1$};
  \node[clause, below=0.3cm of x2] (c2) {$c_2$};

  \draw[send] (x1) to (c1) ;
  \draw[send] (x2) to (c1) ;

  \draw[send] (nx1) to (c2) ;
  \draw[send] (nx2) to (c2) ;
\end{tikzpicture}
\end{center}
      \caption{}
\label{fig:model_a}
\end{subfigure}
    \begin{subfigure}[t]{0.45\textwidth}
      \begin{center}
\begin{tikzpicture}[node distance=1cm, auto,]
  \node[lit] (x1) {$x_1$};
  \node[lit, right=1.0cm of x1] (nx1) {$\flip{x_1}$};
  \node[lit, right=1.5cm of nx1] (x2) {$x_2$};
  \node[lit, right=1.0cm of x2] (nx2) {$\flip{x_2}$};
  \node[clause, below=0.3cm of nx1] (c1) {$c_1$};
  \node[clause, below=0.3cm of x2] (c2) {$c_2$};

  \draw[send] (c1) to (x1) ;
  \draw[send] (c1) to (x2) ;

  \draw[send] (c2) to (nx1) ;
  \draw[send] (c2) to (nx2) ;

  \draw[sendll] (x1) to (nx1);
  \draw[sendll] (nx1) to (x1);
  \draw[sendll] (x2) to (nx2);
  \draw[sendll] (nx2) to (x2);

\end{tikzpicture}
      \end{center}
\caption{}
\label{fig:model_b}
\end{subfigure}
\caption{High-level illustration of NeuroSAT operating on the graph
  representation of $\sat{1|2,\flip{1}|\flip{2}}$.  On the top of
  both figures are nodes for each of the four literals, and on the
  bottom are nodes for each of the two clauses. At every time step $t$,
  we have an embedding for every literal and every clause.  An
  iteration consists of two stages. First, each clause receives messages
  from its neighboring literals and updates it embedding accordingly
  (Figure~\ref{fig:model_a}). Next, each literal receives messages
  from its neighboring clause as well as from its complement, and
  updates its embedding accordingly (Figure~\ref{fig:model_b}).}
\label{fig:model}
\end{center}
\end{figure}
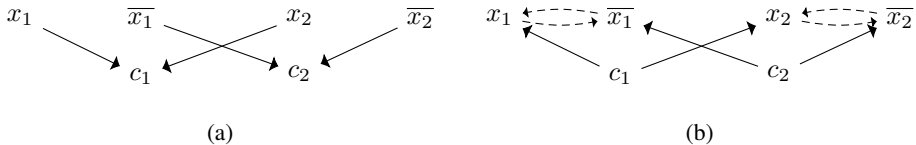

We now describe our neural network architecture, NeuroSAT, that
enforces both permutation invariance and negation invariance.  We
encode a SAT problem as an undirected graph with one node for every
literal, one node for every clause, an edge between every literal and
every clause it appears in, and a different type of edge between each
pair of complementary literals (\eg between $x_i$ and
$\flip{x_i}$). NeuroSAT iteratively refines a vector space embedding
for each node by passing ``messages'' back and forth along the edges
of the graph as described in~\citet{gilmer2017neural}. At every time
step, we have an embedding for every literal and every clause.  An
iteration consists of two stages. First, each clause receives messages
from its neighboring literals and updates its embedding
accordingly. Next, each literal receives messages from its neighboring
clauses as well as from its complement, then updates its embedding
accordingly.  Figure~\ref{fig:model} provides a high-level
illustration of the architecture.

More formally, our model is parameterized by two vectors
($\uscore{L}{init}$, $\uscore{C}{init}$), three multilayer perceptrons
($\uscore{L}{msg}$, $\uscore{C}{msg}$, $\uscore{L}{vote}$) and two
layer-norm LSTMs~\citep{ba2016layer, hochreiter1997lstm}
($\uscore{L}{u}$, $\uscore{C}{u}$).  At every time step $t$, we have a
matrix $L^{(t)} \in \mathbb{R}^{2n \times d}$ whose $i$th row contains
the embedding for the literal $\ell_i$ and a matrix $C^{(t)} \in
\mathbb{R}^{m \times d}$ whose $j$th row contains the embedding for
the clause $c_j$, which we initialize by tiling $\uscore{L}{init}$ and
$\uscore{C}{init}$ respectively.  We also have hidden states
$L_h^{(t)} \in \mathbb{R}^{2n\times d}$ and $C_h^{(t)} \in
\mathbb{R}^{m\times d}$ for $\uscore{L}{u}$ and $\uscore{C}{u}$
respectively, both initialized to zero matrices.  Let $M$ be the
(bipartite) adjacency matrix defined by $M(i, j) = \ind{\ell_i \in
  c_j}$ and let $\Flip$ be the operator that takes a matrix $L$ and
swaps each row of $L$ with the row corresponding to the literal's
negation. A single iteration consists of applying the following
two updates:
\begin{align*}
(C^{(t+1)}, C_h^{(t+1)}) & \gets \uscore{C}{u}( [ C_h^{(t)}, M^\top \uscore{L}{msg}(L^{(t)}) ] ) \\
  (L^{(t+1)}, L_h^{(t+1)}) & \gets \uscore{L}{u}([ L_h^{(t)}, \Flip(L^{(t)}), M \uscore{C}{msg}(C^{(t+1)}) ])
\end{align*}
After $T$ iterations, we compute $L_*^{(T)} \gets
\uscore{L}{vote}(L^{(T)}) \in \mathbb{R}^{2n}$, which contains a
single scalar for each literal (the literal's
\emph{vote}), and then we compute the average of the literal votes $y^{(T)}
\gets \mean(L_*^{(T)}) \in \mathbb{R}$.  We train the network to
minimize the sigmoid cross-entropy loss between the logit $y^{(T)}$
and the true label $\phi(P)$.

Our architecture enforces permutation invariance by
operating on nodes and edges according to the topology of the graph
without any additional ordering over nodes or edges. Likewise, it
enforces negation invariance by treating all literals the same no
matter whether they originated as a positive or negative occurrence of
a variable.

We stress that none of the learned parameters depend on the size of
the SAT problem and that a single model can be trained and tested on
problems of arbitrary and varying sizes. At both train and test time,
the input to the model is simply any bipartite adjacency matrix $M$
over any number of literals and clauses.  The learned parameters only
determine how each individual literal and clause behaves in terms of
its neighbors in the graph.  Variation in problem size is handled by
the aggregation operators: we sum the outgoing messages of each of a
node's neighbors to form the incoming message, and we take the mean of
the literal votes at the end of message passing to form the logit $y^{(T)}$.

\section{Training data}
\label{sec:training_data}

We want our neural network to be able to classify (and ultimately
solve) SAT problems from a variety of domains that it never trained
on. One can easily construct distributions over SAT problems for which
it would be possible to predict satisfiability with perfect accuracy
based only on crude statistics; however, a neural network trained on
such a distribution would be unlikely to generalize to
problems from other domains.  To force our network to learn something
substantive, we create a distribution $\SR(n)$ over pairs of random
SAT problems on $n$ variables with the following property: one element
of the pair is satisfiable, the other is unsatisfiable, and the two
differ by negating only a single literal occurrence in a single
clause.  To generate a random clause on $n$ variables,
$\SR(n)$ first samples a small integer $k$ (with mean a little over 4)
\footnote{We use $1 + \mathbf{Bernoulli}(0.7) + \mathbf{Geo}(0.4)$ so
  that we generate clauses of varying size but with only a small
  number of clauses of length $2$, since too many random clauses of
  length $2$ make the problems too easy on average.} then samples $k$
variables uniformly at random without replacement, and finally negates
each one with independent probability 50\%. It continues to generate
clauses $c_i$ in this fashion, adding them to the SAT problem, and
then querying a traditional SAT solver (we used
Minisat~\cite{sorensson2005minisat}), until adding the clause $c_{m}$
finally makes the problem unsatisfiable. Since $\sat{c_1,\dotsc,c_{m-1}}$ had
a satisfying assignment, negating a single literal in $c_m$ must yield
a satisfiable problem $\sat{c_1,\dotsc,c_{m-1},c_{m}'}$.  The pair
$\left( \sat{c_1,\dotsc,c_{m-1},c_m}, \sat{c_1,\dotsc,c_{m-1},c_m'}
\right)$ are a sample from $\SR(n)$.

\begin{figure*}
  \centering
  \includegraphics[height=3.82cm]{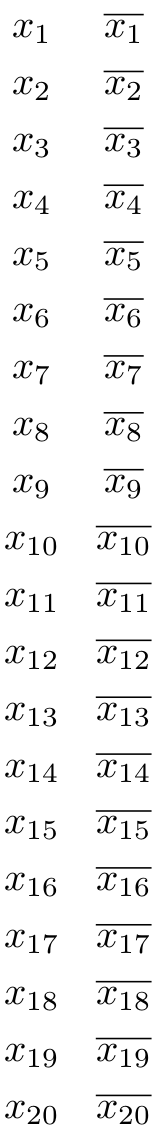}
\forloop{inum}{1}{\value{inum} < 25}{
  \includegraphics[height=3.8cm]{figures/run3022805014702275039_problem=data_dir=simple_n20_npb=0_nb=200_nr=40_rand=0_seed=0_t=1.pkl_v60_axis0_dpi10/round_t\arabic{inum}.png}
}
\\ Iteration $\longrightarrow$
  \caption{The sequence of literal votes $L_*^{(1)}$ to $L_*^{(24)}$
    as NeuroSAT runs on a satisfiable problem from $\SR(20)$.  For
    clarity, we reshape each $L_*^{(t)}$ to be an
    $\mathbb{R}^{n\times2}$ matrix so that each literal is paired with
    its complement; specifically, the $i$th row contains the scalar
    votes for $x_i$ and $\flip{x_i}$.  Here white represents zero,
    blue negative and red positive. For several iterations, almost
    every literal is voting \emph{unsat} with low confidence (light
    blue). Then a few scattered literals start voting \emph{sat} for
    the next few iterations, but not enough to affect the mean
    vote. Suddenly there is a phase transition and all the literals
    (and hence the network as a whole) start to vote \emph{sat} with
    very high confidence (dark red). After the phase transition,
    the vote for each literal converges and the network stops
    evolving.}
\label{fig:animation1}
\end{figure*}

\section{Predicting satisfiability}
\label{sec:predicting_sat}
Although our ultimate goal is to solve SAT problems arising from a
variety of domains, we begin by training NeuroSAT as a classifier to
predict satisfiability on $\SR(40)$.  Problems in $\SR(40)$ are small
enough to be solved efficiently by modern SAT solvers---a fact we rely
on to generate the problems---but the classification problem is highly
non-trivial from a machine learning perspective. Each problem has 40 variables and almost 200
clauses on average, and the positive and negative examples differ by
negating only a single literal occurrence out of a thousand.
We were unable to
train an LSTM on a many-hot encoding of clauses (specialized to
problems with 40 variables) to predict with \textgreater50\% accuracy
on its training set. Even the
canonical SAT solver MiniSAT~\citep{sorensson2005minisat} needs to
backjump\footnote{\ie backtrack multiple steps at a time} almost ten
times on average, and needs to perform over a hundred primitive
logical inferences (\ie unit propagations) to solve each problem.

We instantiated the NeuroSAT architecture described in
\S\ref{sec:model} with $d=128$ dimensions for the literal embeddings,
the clause embeddings, and all the hidden units; $3$ hidden layers and
a linear output layer for each of the MLPs $\uscore{L}{msg}$,
$\uscore{C}{msg}$, and $\uscore{L}{vote}$; and rectified linear units
for all non-linearities. We regularized by the $\ell_2$ norm of the
parameters scaled by $10^{-10}$, and performed $T=26$ iterations of
message passing on every problem. We trained our model using the ADAM
optimizer~\citep{kingma2014adam} with a learning rate of $2 \times
10^{-5}$, clipping the gradients by global norm with clipping ratio
$0.65$~\citep{pascanu2012understanding}. We batched multiple problems
together, with each batch containing up to 12,000 nodes (\ie literals
plus clauses). To accelerate the learning, we sampled the number of
variables $n$ uniformly from between 10 and 40 during training (\ie we
trained on $\SR(\U(10, 40))$), though we only evaluate on
$\SR(40)$. We trained on millions of problems.

After training, NeuroSAT is able to classify the test set correctly
with 85\% accuracy. In the next section, we examine how NeuroSAT
manages to do so and show how we can decode solutions to satisfiable
problems from its activations.  Note: for the entire rest of the
paper, \emph{NeuroSAT} refers to the specific trained model that has
only been trained on $\SR(\U(10, 40))$.

\section{Decoding satisfying assignments}
\label{sec:decode}

Let us try to understand what NeuroSAT (trained on $\SR(\U(10, 40))$) is computing as it runs on new problems at test time.
For a given run, we can compute and visualize the $2n$-dimensional vector of literal
votes $L_*^{(t)} \gets \uscore{L}{vote}(L^{(t)})$ at
every iteration $t$.
Figure~\ref{fig:animation1} illustrates the sequence of
literal votes $L_*^{(1)}$ to $L_*^{(24)}$ as NeuroSAT runs on a
satisfiable problem from $\SR(20)$.  For clarity, we reshape each
$L_*^{(t)}$ to be an $\mathbb{R}^{n\times2}$ matrix so that each
literal is paired with its complement; specifically, the $i$th row
contains the scalar votes for $x_i$ and $\flip{x_i}$.
Here white represents zero, blue negative and red positive. For
several iterations, almost every literal is voting \emph{unsat} with
low confidence (light blue). Then a few scattered literals start
voting \emph{sat} for the next few iterations, but not enough to
affect the mean vote. Suddenly, there is a phase transition and all
the literals (and hence the network as a whole) start to vote
\emph{sat} with very high confidence (dark red). After the phase transition,
the vote for each literal converges and the network stops evolving.

NeuroSAT seems to exhibit qualitatively similar behavior on every
satisfiable problem that it predicts
correctly. The
problems for which NeuroSAT guesses \emph{unsat} are similar except
without the phase change: it continues to guess \emph{unsat} with
low-confidence for as many iterations as NeuroSAT runs for.  NeuroSAT
never becomes highly confident that a problem is \emph{unsat}, and it
almost never guesses \emph{sat} on an \emph{unsat} problem. These
results suggest that NeuroSAT searches for a certificate of
satisfiability, and that it only guesses \emph{sat} once it has found
one.

\begin{figure}[ht]
  \centering
    \fbox{\includegraphics[width=0.11\textwidth]{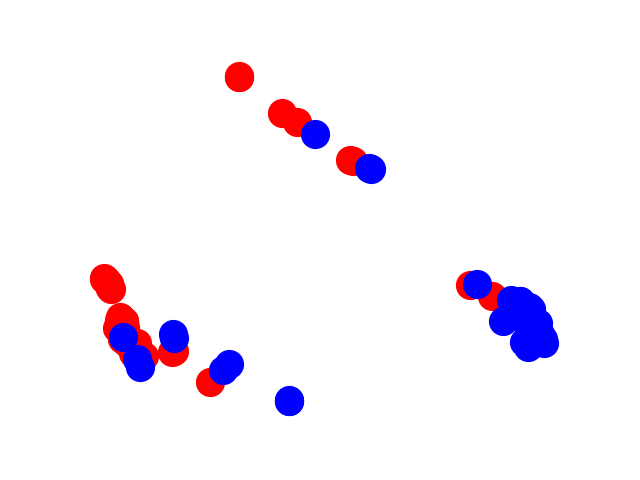}}
    \fbox{\includegraphics[width=0.11\textwidth]{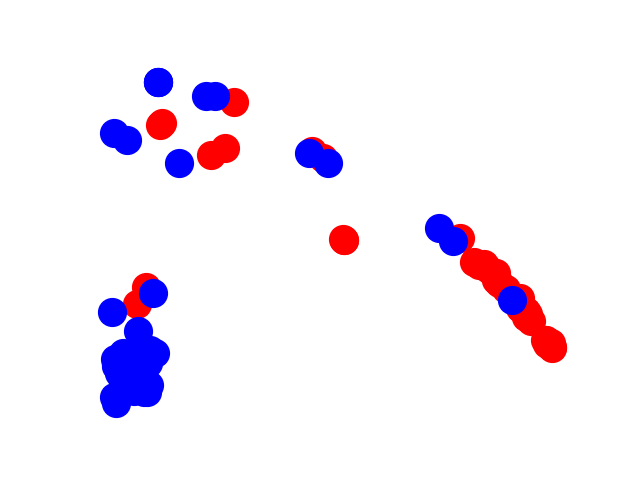}}
    \fbox{\includegraphics[width=0.11\textwidth]{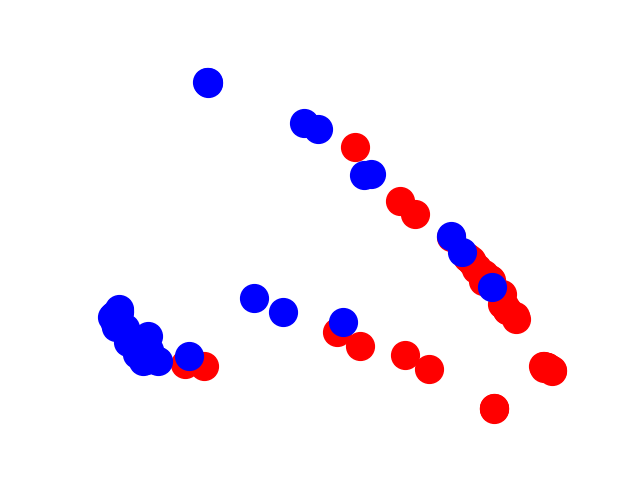}}
    \fbox{\includegraphics[width=0.11\textwidth]{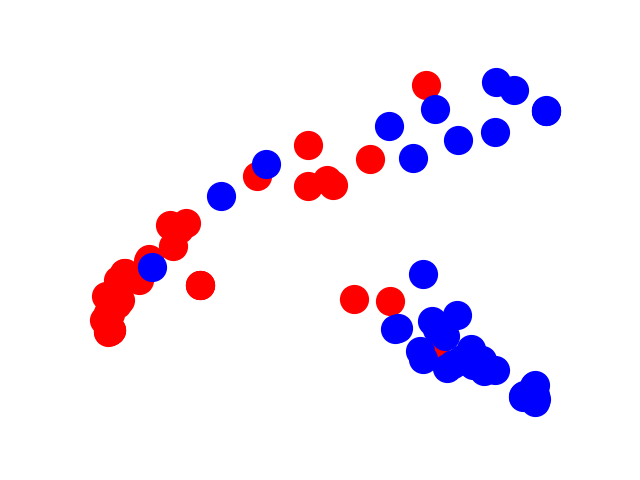}}
    \fbox{\includegraphics[width=0.11\textwidth]{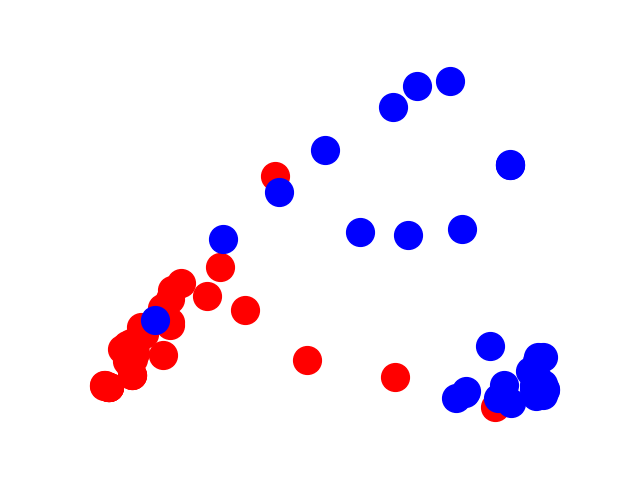}}
    \fbox{\includegraphics[width=0.11\textwidth]{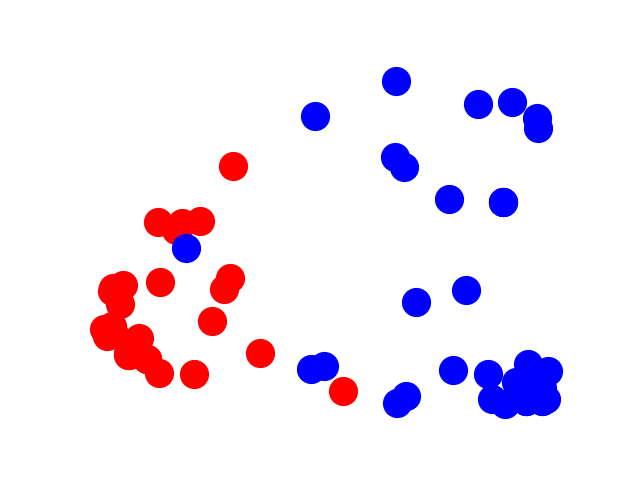}}
    \fbox{\includegraphics[width=0.11\textwidth]{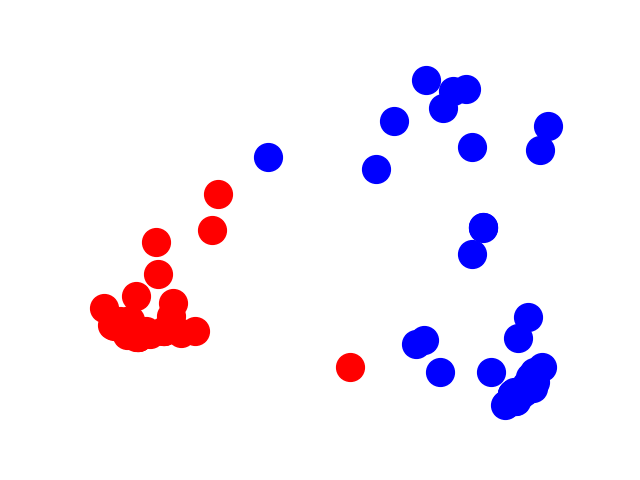}}
    \fbox{\includegraphics[width=0.11\textwidth]{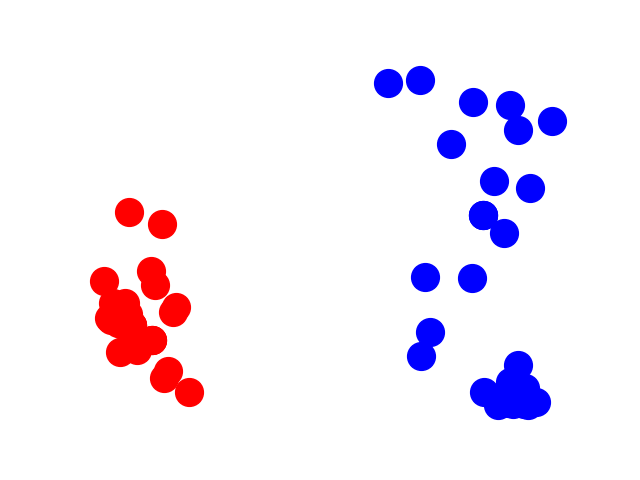}} \\
    Iteration $\longrightarrow$
    \caption{PCA projections for the high-dimensional literal
      embeddings $L^{(12)}$ to $L^{(26)}$ (skipping every other time
      step) as NeuroSAT runs on a satisfiable problem from $\SR(40)$.
      Blue and red dots indicate literals that are set to 0 and 1 in
      the satisfying assignment that it eventually finds,
      respectively. We see that the blue and red dots are mixed up and
      cannot be linearly separated until the phase transition at the
      end, at which point they form two distinct clusters according to
      the satisfying assignment.}
    \label{fig:pca}
\end{figure}

Let us look more carefully at the literal votes $L_*^{(24)}$ from
Figure~\ref{fig:animation1} after convergence.  Note that most of the
variables have one literal vote distinctly darker than the
other. Moreover, the dark votes are all approximately equal to each
other, and the light votes are all approximately equal to each other
as well. Thus the votes seem to encode one bit for each variable.  It
turns out that these bits encode a satisfying assignment in this case,
but they do not do so reliably in general.  Recall from
\S\ref{sec:model} that NeuroSAT projects the higher dimensional
literal embeddings $L^{(T)} \in \mathbb{R}^{2n\times d}$ to the
literal votes $L_*^{(T)}$ using the MLP $\uscore{L}{vote}$.
Figure~\ref{fig:pca} illustrates the two-dimensional PCA embeddings
for $L^{(12)}$ to $L^{(26)}$ (skipping every other time step) as
NeuroSAT runs on a satisfiable problem from $\SR(40)$. Blue and red
dots indicate literals that are set to 0 and 1 in the satisfying
assignment that it eventually finds, respectively. The blue and red
dots cannot be linearly separated until the phase transition at the
end, at which point they form two distinct clusters according to the
satisfying assignment. We observe a similar clustering almost every
time the network guesses \emph{sat}.  Thus the literal votes
$L_*^{(T)}$ only ever encode the satisfying assignment by chance, when
the projection $\uscore{L}{vote}$ happens to preserve this clustering.

Our analysis suggests a more reliable
way to decode solutions from NeuroSAT's internal activations:
2-cluster $L^{(T)}$ to get cluster centers $\Delta_1$ and $\Delta_2$,
partition the variables according to the predicate
\(
\| x_i - \Delta_1 \|^2 + \| \flip{x_i} - \Delta_2 \|^2 < \| x_i - \Delta_2 \|^2 + \| \flip{x_i} - \Delta_1 \|^2
\),
and then try both candidate assignments that result from mapping the
partitions to truth values.
This decoding procedure (using $k$-means to find the two cluster centers) successfully decodes a satisfying assignment
for over $70\%$ of the satisfiable problems in the $\SR(40)$ test set.
Table~\ref{table:summary_sr40} summarizes the results when training on
$\SR(\U(10, 40))$ and testing on $\SR(40)$.

\setlength{\fboxsep}{0pt}


\begin{table}
  \small
  \begin{center}
  \begin{tabular}{|rl|}
    \hline
    Trained on: & $\SR(\U(10, 40))$ \\
    Trained with: & 26 iterations \\
    Tested on: & $\SR(40)$ \\
    Tested with: & 26 iterations \\
    Overall test accuracy: & 85\% \\
    Accuracy on \emph{unsat} problems: & 96\% \\
    Accuracy on \emph{sat} problems:   & 73\% \\
    \textbf{Percent of \emph{sat} problems solved:} & \textbf{70\%} \\
  \hline
\end{tabular}
\end{center}
\caption{NeuroSAT's performance at test time on $\SR(40)$ after
  training on $\SR(\U(10, 40))$.  It almost never guesses \emph{sat}
  on unsatisfiable problems. On satisfiable problems, it correctly
  guesses \emph{sat} 73\% of the time, and we can decode a satisfying
  assignment for 70\% of the satisfiable problems by clustering the
  literal embeddings $L^{(T)}$ as described in \S\ref{sec:decode}.}
\label{table:summary_sr40}
\end{table}
Recall that at training time, NeuroSAT is only given \emph{a single
  bit} of supervision for each SAT problem. Moreover, the positive and
negative examples in the dataset differ only by the placement of a
single edge. NeuroSAT has learned to search for satisfying assignments
solely to explain that single bit of supervision.

\section{Extrapolating to other problem distributions}
\label{sec:generalize}
\subsection{Bigger problems}

\begin{figure}[ht]
  \centering
  \begin{minipage}{0.48\textwidth}
    \centering
      \includegraphics[height=130pt]{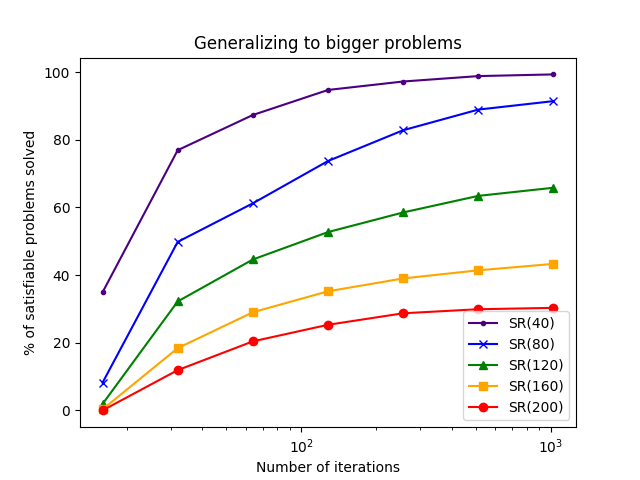}
    \caption{ NeuroSAT's success rate on $\SR(n)$ for a range of $n$ as a
      function of the number of iterations $T$. Even though we only train
      NeuroSAT on $\SR(40)$ and below, it is able to solve SAT problems
      sampled from $\SR(n)$ for $n$ much larger than $40$ by simply
      running for more iterations.}
    \label{fig:bigger}
  \end{minipage}
  \hspace{10pt}
  \begin{minipage}{0.48\textwidth}
    \centering
      \includegraphics[height=130pt]{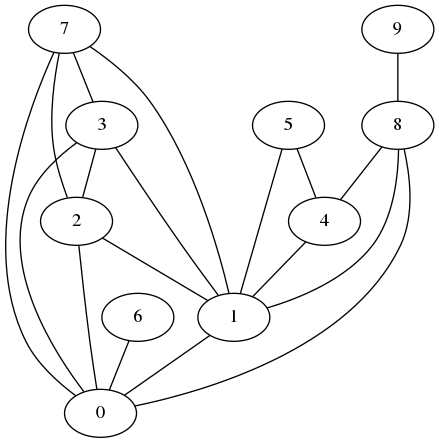}
    \caption{Example graph from the Forest-Fire distribution.  The graph
      has a coloring for $k \geq 5$, a clique for $k \leq 3$, a dominating
      set for $k \geq 3$, and a vertex cover for $k \geq 6$.  However,
      these properties are not perceptually obvious and require deliberate
      computation to determine.}
    \label{fig:graph_vis}
  \end{minipage}
\end{figure}

Even though we only train NeuroSAT on $\SR(\U(10, 40))$, it is able
to solve SAT problems sampled from $\SR(n)$ for $n$ much larger than
$40$ by simply running for more iterations of message passing.  Figure~\ref{fig:bigger}
shows NeuroSAT's success rate on $\SR(n)$ for a range of $n$ as a
function of the number of iterations $T$. For $n=200$, there are
$2^{160}$ times more possible assignments to the variables than any problem
it saw during training, and yet it can solve 25\% of the satisfiable
problems in $\SR(200)$ by running for four times more iterations than
it performed during training.  On the other hand, when restricted
to the number of iterations it was trained with, it solves under 10\%
of them. Thus we see that its ability to solve bigger and harder
problems depends on the fact that the dynamical system it
has learned encodes generic procedural knowledge that can operate
effectively over a wide range of time frames.

\subsection{Different problems}

Every problem in \textbf{NP} can be reduced to SAT in polynomial time,
and SAT problems arising from different domains may have radically
different structural and statistical properties. Even though NeuroSAT
has learned to search for satisfying assignments on problems from
$\SR(n)$, we may still find that the dynamical system it has learned
only works properly on problems similar to those it was trained on.

To assess NeuroSAT's ability to extrapolate to different classes of
problems, we generated problems in several other domains and then
encoded them all into SAT problems (using standard encodings).  In
particular, we started by generating one hundred graphs from each of
six different random graph distributions (Barabasi, Erd\"{o}s-Renyi,
Forest-Fire, Random-$k$-Regular, Random-Static-Power-Law, and
Random-Geometric).\footnote{See~\citet{newman2010networks} for an
  overview of random graph distributions.}  We found parameters for
the random graph generators such that each graph has ten nodes and
seventeen edges on average. For each graph in each collection, we
generated graph coloring problems ($3 \leq k \leq 5$), dominating-set
problems ($2 \leq k \leq 4)$), clique-detection problems ($3 \leq k
\leq 5$), and vertex cover problems ($4 \leq k \leq
6$).\footnote{See~\citep{lewis1983computers} for an overview of these
  problems as well as the standard encodings.} We chose the range of
$k$ for each problem to include the threshold for most of the graphs
while avoiding trivial problems such as $2$-clique. As before, we used
Minisat~\cite{sorensson2005minisat} to determine satisfiability.
Figure~\ref{fig:graph_vis} shows an example graph from the
distribution. Note that the trained network does not know anything
\emph{a priori} about these tasks; the generated SAT problems need to
encode not only the graphs themselves but also formal descriptions of
the tasks to be solved.

Out of the 7,200 generated problems, we kept only the 4,888
satisfiable problems.  On average these problems contained over two
and a half times as many clauses as the problems in
$\SR(40)$. We ran NeuroSAT for 512 iterations on each of them
  and found that we could successfully decode solutions for 85\% of
  them. In contrast, Survey Propagation (SP)~\citep{braunstein2005survey}, the canonical (learning-free)
message passing algorithm for satisfiability, does not on its own
converge to a satisfying assignment on any of these
problems.\footnote{We implementated the version with reinforcement
  messages described in \citet{knuth2015sat}, along with the numerical
  trick explained in Exercise 359.}  This suggests that NeuroSAT has
not simply found a way to approximate SP, but rather has
synthesized a qualitatively different algorithm.

\section{Finding unsat cores}
\label{sec:unsat}

NeuroSAT (trained on $\SR(\U(10, 40))$) can find satisfying
assignments but is not helpful in constructing proofs of
unsatisfiability.  When it runs on an unsatisfiable problem, it keeps
searching for a satisfying assignment indefinitely and
non-systematically.  However, when we train the same architecture on a
dataset in which each unsatisfiable problem has a small subset of
clauses that are already unsatisfiable (called an \emph{unsat core}),
it learns to detect these unsat cores instead of searching for
satisfying assignments.  The literals involved in the unsat core can
be decoded from its internal activations. When the number of literals
involved in the unsat core is small relative to the total number of
literals, knowing the literals involved in the unsat core can enable
constructing a resolution proof more efficiently.

We generated a new distribution $\SRC(n, u)$ that is similar to
$\SR(n)$ except that every unsatisfiable problem contains a small
unsat core.  Here $n$ is the number of variables as before, and $u$ is
an unsat core over $x_1, \dotsc, x_k$ ($k < n$) that can be made into
a satisfiable set of clauses $u'$ by negating a single literal. We
sample a pair from $\SRC(n, u)$ as follows.  First, we initialize a
problem with $u'$, and then we sample clauses (over $x_1$ to $x_n$)
just as we did for $\SR(n)$ until the problem becomes
unsatisfiable. We can now negate a literal in the final clause to get
a satisfiable problem $p_s$, and then we can swap $u'$ for $u$ in
$p_s$ to get $p_u$, which is unsatisfiable since it contains the unsat
core $u$.  We created train and test datasets from $\SRC(40, u)$ with
$u$ sampled at random for each problem from a collection of three
unsat cores ranging from three clauses to nine clauses: the unsat core
$R$ from~\citet{knuth2015sat}, and the two unsat cores resulting from
encoding the pigeonhole principles $\PP(2, 1)$ and $\PP(3, 2)$.\footnote{The pigeonhole principle and the standard SAT encoding are described in~\citet{knuth2015sat}.}  We
trained our architecture on this dataset, and we refer to the trained
model as \emph{NeuroUNSAT}.

\begin{figure}[ht]
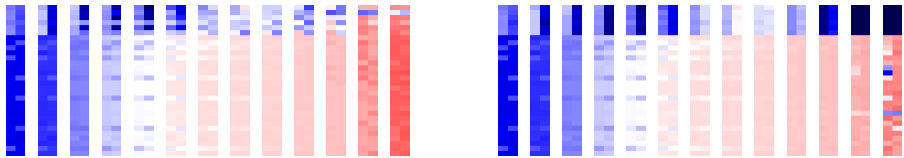

\centering
  \begin{subfigure}[t]{0.45\textwidth}
\begin{center}
\forloop{inum}{1}{\value{inum} < 14}{
 \includegraphics[height=2.0cm, width=0.25cm]{figures/run3671678704658005230_20_problem=data_dir=pp_3_2_n30_npb=1000000_nb=20_nr=14_rand=0_seed=0_t=11.pkl_v60_axis0_dpi10/round_t\arabic{inum}.png}
}
\end{center}
\caption{NeuroUNSAT running on a satisfiable problem.}
\label{fig:unsat_animation_sat}
\end{subfigure}\hspace{5pt}
\begin{subfigure}[t]{0.45\textwidth}
  \begin{center}
\forloop{inum}{1}{\value{inum} < 14}{
 \includegraphics[height=2.0cm, width=0.25cm]{figures/run3671678704658005230_20_problem=data_dir=pp_3_2_n30_npb=1000000_nb=20_nr=14_rand=0_seed=0_t=10.pkl_v60_axis0_dpi10/round_t\arabic{inum}.png}
}
\end{center}

\caption{NeuroUNSAT running on an unsatisfiable problem.}
\label{fig:unsat_animation_unsat}

\end{subfigure}

  \caption{The sequence of literal votes $L_*^{(t)}$ as NeuroUNSAT
    runs on a pair of problems from $\SRC(30, \PP(3, 2))$. In both
    cases, the literals in the first six rows are involved in the
    unsat core. In \ref{fig:unsat_animation_sat}, NeuroUNSAT inspects the modified
    core $u'$ of the satisfiable problem but concludes that it does
    not match the pattern. In \ref{fig:unsat_animation_unsat},
    NeuroUNSAT finds the unsat core $u$ and
    votes \emph{unsat} with high confidence (dark blue).}
  \label{fig:unsat_animations}
\end{figure}

NeuroUNSAT is able to predict satisfiability on the test set with
100\% accuracy. Upon inspection, it seems to do so by learning to
recognize the unsat cores.  Figure~\ref{fig:unsat_animations} shows
NeuroUNSAT running on a pair of problems from $\SRC(30, \PP(3,
2))$. In both cases, the literals in the first six rows are involved
in the unsat core. In Figure~\ref{fig:unsat_animation_sat}, NeuroUNSAT
inspects the modified core $u'$ of the satisfiable problem but
concludes that it does not match the pattern exactly. In
Figure~\ref{fig:unsat_animation_unsat}, NeuroUNSAT finds the unsat
core $u$ and votes \emph{unsat} with high confidence (dark blue). As in
\S\ref{sec:decode}, the literals involved in the unsat core can
sometimes be decoded from the literal votes $L_*^{(T)}$, but it is
more reliable to 2-cluster the higher-dimensional literal embeddings
$L^{(T)}$. On the test set, the small number of literals involved in
the unsat core end up in their own cluster 98\% of the time.

Note that we do not expect NeuroUNSAT to generalize to arbitary unsat
cores: as far as we know it is simply memorizing a collection of
specific subgraphs, and there is no evidence it has learned a generic
procedure to prove \emph{unsat}.

\section{Related work}

There have been many attempts over the years to apply statistical
learning to various aspects of the SAT problem: restart
strategies~\citep{haim2009restart}, branching
heuristics~\citep{liang2016learning, grozea2014can, flint2012perceptron}, parameter
tuning~\citep{singh2009avatarsat}, and solver
selection~\citep{xu2008satzilla}. None of these approaches use neural
networks, and instead make use of both generic graph features and
features extracted from the runs of SAT solvers. Moreover, these
approaches are designed to assist existing solvers and do not aim to
solve SAT problems on their own.

From the machine learning perspective, the closest work to ours is
\citet{palm2017recurrent}, which showed that an MPNN can be trained to
predict the unique solutions of Sudoku puzzles. We believe that their
network's success is an instance of the phenomenon we study in this
paper, namely that MPNNs can synthesize local search algorithms for
constraint satisfaction problems. \citet{evans2018can} present a
neural network architecture that can learn to predict whether one
propositional formula entails another by randomly sampling and
evaluating candidate assignments. Unlike NeuroSAT, their network does
not perform heuristic search and can only work on simple problems for
which random guessing is tractable.  There have also been several
recent papers showing that various neural network architectures can
learn good heuristics for $\myNP$-hard combinatorial optimization
problems~\citep{vinyals2015pointer, bello2016neural, dai2017learning};
however, finding low-cost solutions to optimization problems requires
less precise reasoning than finding satisfying assignments.

\section{Discussion}

Our main motivation has been scientific: to better understand the
extent to which neural networks are capable of precise, logical
reasoning.  Our work has definitively established that neural networks
can learn to perform discrete search on their own without the help of
hard-coded search procedures, even after only end-to-end training with
minimal supervision.  We found this result surprising and think it
constitutes an important contribution to the community's evolving
understanding of the capabilities and limitations of neural networks.

Although not our primary concern, we also hope that our findings
eventually lead to improvements in practical SAT solving. As we
stressed early on, as an end-to-end SAT solver the trained NeuroSAT
system discussed in this paper is still vastly less reliable than the
state-of-the-art. We concede that we see no obvious path to beating
existing SAT solvers.  One approach might be to continue to train
NeuroSAT as an end-to-end solver on increasingly difficult problems. A
second approach might be to use a system like NeuroSAT to help guide
decisions within a more traditional SAT solver, though it is not clear
that NeuroSAT provides any useful information before it finds a
satisfying assignment. However, as we discussed in \S\ref{sec:unsat},
when we trained our architecture on different data it learned an
entirely different procedure. In a separate experiment omitted for
space reasons, we also trained our architecture to predict whether
there is a satisfying assignment involving each individual literal in
the problem and found that it was able to predict these bits with high
accuracy as well. Unlike NeuroSAT, it made both type I and type II
errors, had no discernable phase transition, and could make reasonable
predictions within only a few rounds. We believe that architectures
descended from NeuroSAT will be able to learn very different
mechanisms and heuristics depending on the data they are trained on
and the details of their objective functions.  We are cautiously
optimistic that a descendant of NeuroSAT will one day lead to
improvements to the state-of-the-art.

\subsubsection*{Acknowledgements}
We thank Steve Mussmann, Alexander Ratner, Nathaniel Thomas, Vatsal
Sharan and Cristina White for providing valuable feedback on early
drafts. We also thank William Hamilton, Geoffrey Irving and Arun
Chaganty for helpful discussions. This work was supported by Future of
Life Institute grant 2017-158712.

\bibliography{main}
\bibliographystyle{iclr2019_conference}

\end{document}